\documentclass[runningheads]{llncs}
\usepackage[T1]{fontenc}
\usepackage{todonotes}
\usepackage{graphicx}
\usepackage{booktabs}
\usepackage[misc]{ifsym}
\usepackage{amssymb}
\usepackage{amsmath}
\usepackage{CJKutf8}
\usepackage{CJK}
\usepackage[utf8]{inputenc} 
\usepackage[T1]{fontenc}    
\usepackage{url}            
\usepackage{booktabs}       
\usepackage{amsfonts}       
\usepackage{nicefrac}       
\usepackage{microtype}      
\usepackage{lipsum}
\usepackage{pifont}
\usepackage[inline]{enumitem}
\usepackage{boldline}
\usepackage{hhline}
\usepackage{multirow}
\usepackage{xcolor}
\usepackage{fontawesome5}
\usepackage{tablefootnote} 
\usepackage{listings}
\usepackage{booktabs}
\usepackage[table]{xcolor} 
\newcommand{\cmark}{\textcolor{green!60!black}{$\checkmark$}}
\newcommand{\xmark}{\textcolor{red!70!black}{$\times$}}
\usepackage[colorlinks=true, linkcolor=blue, citecolor=blue, urlcolor=blue]{hyperref}

\usepackage{mwe}

\begin{document}

\title{Task-Differentiated Atomic Skill Expansion and Routing for Continual Learning Across Highly Heterogeneous Tasks}

\titlerunning{Task-Differentiated Atomic Skill Expansion and Routing}

\author{Jiacheng Wang\inst{1} \and
Xinjia He\inst{2} \and 
Qi Ding\inst{1} \and 
Yutao Yang\inst{1} \and 
Jie Zhou\inst{1}\textsuperscript{(\Letter)} \and
Liyang Yu\inst{1,3} \and 
Liang Dou\inst{1} \and 
Qin Chen\inst{1}}

\authorrunning{J. Wang et al.}

\tocauthor{Jiacheng Wang, Xinjia He, Qi Ding, Yutao Yang, Jie Zhou, Liyang Yu, Liang Dou, Qin Chen}
\toctitle{Task-Differentiated Atomic Skill Expansion and Routing for Continual Learning Across Highly Heterogeneous Tasks}

\institute{East China Normal University \\ 
\email{jzhou@cs.ecnu.edu.cn}
\and
Fudan University
\and
Ocean University of China
}

\maketitle              

\begin{abstract}
Continual learning (CL) is commonly studied under the assumption that sequential tasks are semantically related or structurally similar. However, in highly heterogeneous settings, where tasks differ substantially in reasoning patterns and input-output formats, existing methods often suffer from catastrophic forgetting and inefficient capacity allocation. To address this challenge, we propose Task-differentiated Atomic Skill Expansion and Routing (\texttt{TASER}), a CL framework that jointly determines how many new atomic skills to introduce for each task and which skills to activate. The framework first uses atomic skill incremental learning to dynamically expand capacity based on task divergence and model uncertainty. It then applies orthogonality-enhanced skill detection to ensure these skills remain semantically distinct and independently reusable. Finally, a skill dynamic routing mechanism composes task-relevant skills through lightweight task-conditioned gating. We further introduce \texttt{HeteroCLBench}, a highly heterogeneous benchmark for CL, comprising 19 diverse tasks across 9 cognitive dimensions under a standardized sequential protocol. Experiments on \texttt{HeteroCLBench} show that \texttt{TASER} consistently outperforms strong baselines by improving plasticity and reducing catastrophic forgetting.

\end{abstract}

\section{Introduction}
\label{sec:intro}
Continual learning balances plasticity and stability by training models sequentially without full retraining, preserving past performance~\cite{wang2024comprehensive,yang2025recent}. This adaptability is crucial for large language models (LLMs) navigating continuously evolving real-world demands. Yet, most continual learning studies for LLMs assume low task heterogeneity, focusing on structurally or semantically similar tasks where shared mechanisms simplify transfer~\cite{chen2024coin,ding2024boosting,huai2025adaptive}. Practical deployments rarely follow such benign streams; LLMs must frequently shift across highly disparate domains, such as mathematical reasoning, legal summarization, and creative writing, which differ drastically in semantics, objectives, reasoning patterns, and input-output formats.

This disparity highlights the necessity of evaluating continual learning under high task heterogeneity. The core challenge shifts from merely retaining prior knowledge to simultaneously enabling specialization for distinct tasks and reusing shared capabilities. Fixed-capacity models are often inadequate here: preserving old parameters alone cannot accommodate emerging skills, while indiscriminate expansion risks redundancy and interference. This perspective aligns with biological lifelong learning; rather than relying on fixed representations, the human brain manages complex experiences via neuroplasticity and structural resource reallocation, mitigating interference between old and new knowledge~\cite{kumaran2016learning,parisi2019continual}. Therefore, we argue that continual learning under high task heterogeneity demands adaptive, modular capacity growth alongside parameter preservation.

Two primary challenges arise under high task heterogeneity. First, the extreme divergence among heterogeneous tasks causes severe catastrophic forgetting, as updates for new domains strongly disrupt previously learned representations~\cite{rusu2016progressive}. Second, the varying degrees of difference between tasks make the required capacity expansion difficult to determine in advance~\cite{shi2025continual_survey,yoon2017lifelong}. Because tasks require a mixture of shared and specialized skills, it is difficult to determine whether new atomic skills should be added for an incoming task, and if so, exactly how many~\cite{rusu2016progressive,yoon2017lifelong}. Consequently, an effective continual learning framework must dynamically manage atomic skill expansion while isolating task-specific knowledge and enabling flexible cross-task composition.

To address these challenges, we propose Task-differentiated Atomic Skill Expansion and Routing (\texttt{TASER}), a continual learning framework designed for highly heterogeneous task streams. \texttt{TASER} consists of three complementary components. The atomic skill incremental learning component adaptively expands model capacity by introducing new atomic skills when needed. The orthogonality-enhanced skill detection mechanism encourages skill separation and independent reuse. Finally, the skill dynamic routing module composes task-relevant skills through lightweight task-conditioned gating. These components enable \texttt{TASER} to balance task-specific specialization with cross-task skill sharing, improving both stability and plasticity in heterogeneous continual learning.

Our contributions are as follows:
\begin{itemize}[leftmargin=*, align=left]
\item This work formulates continual learning under high task heterogeneity and introduces \texttt{HeteroCLBench}, a benchmark of 19 functionally diverse tasks with a standardized sequential evaluation protocol. 
\item We develop \texttt{TASER} as a modular continual learning framework that jointly performs adaptive atomic skill expansion, skill separation, and dynamic routing for heterogeneous task streams. 
\item Experiments on \texttt{HeteroCLBench} and the LNT benchmark demonstrate that \texttt{TASER} consistently outperforms strong continual learning baselines, reducing forgetting while improving plasticity and transfer.
\end{itemize}

\section{Related Work}
\label{sec:related}

\textbf{Continual Learning.}
Continual learning aims to acquire new knowledge sequentially without catastrophic forgetting~\cite{mccloskey1989catastrophic}. Existing approaches are commonly grouped into three paradigms: \textit{regularization-based methods}, such as EWC~\cite{huszar2018note_ewc} and LwF~\cite{li2017learning_lwf}, which restrict updates to parameters important for previous tasks; \textit{replay-based methods}~\cite{rolnick2019experience_replay}, which retain or regenerate historical data during training; and \textit{subspace-based methods}, which reduce interference by constraining updates to directions less harmful to prior tasks. In particular, OGD~\cite{farajtabar2020orthogonal_ogd} and GPM~\cite{saha2021gradient_gpm} identify protected subspaces from activations via SVD, while O-LoRA~\cite{wang2023orthogonal_olora_LNT} extends this idea to parameter-efficient adaptation by assigning orthogonal low-rank subspaces to sequential tasks. Although effective in mitigating forgetting, these methods generally do not explicitly address continual learning under highly heterogeneous task streams, where task complexity and transferability can vary substantially.

\textbf{Parameter-Efficient Continual Learning.}
PEFT-based continual learning, especially LoRA-style adaptation~\cite{hu2022lora}, has become a practical choice for LLMs due to its low training and storage cost. Recent methods improve knowledge retention through weight interpolation (e.g., I-LoRA~\cite{li2025analyzing}) or expert routing mechanisms (e.g., MoE-LoRA~\cite{chen2024coin}, CL-MoE~\cite{huai2025cl}). However, these approaches typically rely on pre-defined adapter ranks or expert budgets, which may be suboptimal when sequential tasks exhibit large variation in representational demands. Our method instead adapts capacity to task complexity while preserving separation across reusable atomic skills.

\textbf{Benchmarks for Continual Learning.}
Standardized benchmarks are essential for evaluating continual learning paradigms. Early evaluations mainly relied on simple vision datasets (e.g., Split-MNIST, Split-CIFAR) or narrow NLP tasks (e.g., text classification streams). More recent efforts have sought to build broader benchmarks for LLMs. For example, TRACE~\cite{wang2023trace} and CITB~\cite{zhang2023citb} introduced task sequences spanning multiple domains, while the LNT benchmark~\cite{wang2023orthogonal_olora_LNT} further expanded evaluation to 15 distinct NLP tasks. Despite this progress, existing benchmarks still lack a structured taxonomy covering broad cognitive macro-dimensions, and often limit evaluation to only 3 or 4 core capabilities. Their task sequences also tend to exhibit high semantic overlap or identical output modalities. To explicitly assess model plasticity and stability under high task heterogeneity, we introduce \texttt{HeteroCLBench}, spanning 9 rigorously defined cognitive dimensions and providing a more realistic lifelong learning environment.

\begin{figure}[t]
  \centering
  \includegraphics[width=\textwidth]{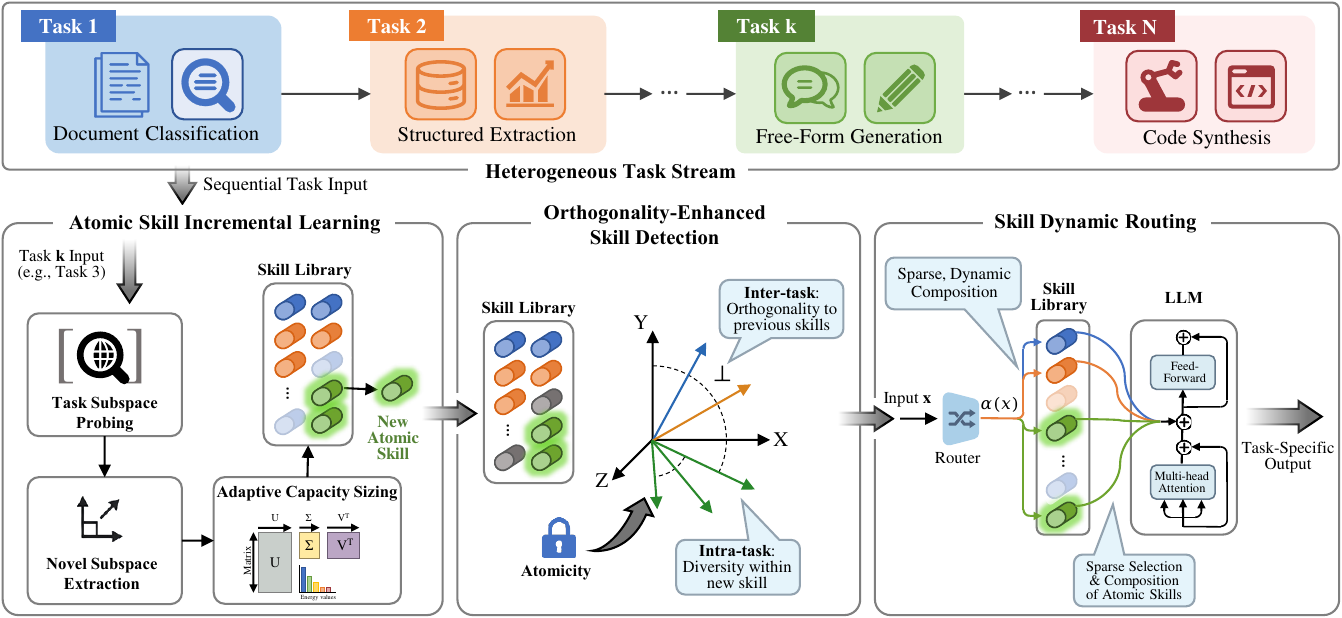} 
  \caption{\textbf{Overview of the \texttt{TASER} Framework.} \texttt{TASER} jointly optimizes skill expansion and routing with three parts: (i) an incremental learning module that dynamically allocates new atomic skills based on task divergence; (ii) an orthogonality-enhanced detection mechanism that enforces functional encapsulation and skill atomicity during training; and (iii) a dynamic routing module that composes active skills for inference via lightweight, task-conditioned gating.}
  \label{fig:framework}
\end{figure}

\section{Method}
\label{sec:method}
To tackle continual learning under \emph{high task heterogeneity}---where sequential tasks exhibit low semantic overlap and divergent input-output structures---we propose the \textbf{T}ask-differentiated \textbf{A}tomic \textbf{S}kill \textbf{E}xpansion and \textbf{R}outing (\texttt{TASER}) framework (Fig.~\ref{fig:framework}). \texttt{TASER} organizes model capacity into a growing library of \emph{atomic skills} (reusable functional modules). It consists of three coordinated components: (1) Atomic Skill Incremental Learning for adaptive capacity expansion; (2) Orthogonality-enhanced Skill Detection to preserve skill atomicity and reduce interference; and (3) Skill Dynamic Routing for task-conditioned skill composition.

Formally, we study a sequential continual learning setting with a highly heterogeneous task stream $\{\mathcal{T}_1,\ldots,\mathcal{T}_N\}$, where each task $\mathcal{T}_k=\langle \mathcal{D}_k,\mathcal{X}_k,\mathcal{Y}_k\rangle$ may differ in data distribution $\mathcal{D}_k$, input space $\mathcal{X}_k$, and output structure $\mathcal{Y}_k$. The goal is to learn $\mathcal{T}_k$ while maintaining performance on all previously observed tasks under this sequential, current-data-only constraint.

\subsection{Atomic Skill Incremental Learning}
\label{sec:asil}

\texttt{TASER} maintains a dynamically growing library of atomic skills implemented via parameter-efficient adapters (e.g., LoRA).
Let the frozen backbone be $\theta_0$.

Before learning task $\mathcal{T}_k$, the library comprises the previously allocated atomic skill modules $\mathcal{S}_{k-1}=\{\theta_1,\dots,\theta_{k-1}\}$, where $\theta_m=\{A^{(m)},B^{(m)}\}$ (layer indices omitted) denotes the LoRA module introduced for task $\mathcal{T}_m$. Each module has rank $r_m$ and encapsulates $r_m$ atomic skills, with each rank dimension treated as one distinct atomic skill; accordingly, the total number of accumulated atomic skills before learning $\mathcal{T}_k$ is $M_{k-1}=\sum_{m=1}^{k-1} r_m$. For notational simplicity, we index atomic skills directly after flattening all previously learned LoRA modules along their rank dimensions.

Under high task heterogeneity, a fixed capacity allocation is often either insufficient for divergent tasks or wasteful for related ones.
Therefore, for each incoming task $\mathcal{T}_k$, \texttt{TASER} dynamically allocates a single new adapter.
The rank of this newly added module---and consequently the exact number of new atomic skills---is determined by estimating the \emph{novel update subspace} induced by $\mathcal{D}_k$ relative to the historical subspace.
This procedure involves three steps.

\paragraph{Step A: Task Subspace Probing.}
We first fit an over-parameterized probe adapter for a few steps to reveal the update directions preferred by the new task.
Concretely, for task $\mathcal{T}_k$, we instantiate a temporary probe LoRA module with rank $r_{\text{probe}}$, while keeping all historical skills in $\mathcal{S}_{k-1}$ frozen.
This probe rank defines the maximum capacity available to the new skill before SVD truncation; therefore, in later analyses, we also denote it as the maximum capacity budget $r_{\text{max}}$.
Importantly, the probe is not required to converge. It only needs to produce a sufficiently stable singular spectrum that captures the dominant task-induced update directions.

\paragraph{Step B: Novel Subspace Extraction.}
Next, we remove directions that are already covered by previous skills, and keep only the residual (new) components.
We form the historical down-projection matrix by concatenating the down projections of all previous skills:
\begin{equation}
A_{\text{hist}} = \mathrm{Concat}\big(A_1,\dots,A_{M_{k-1}}\big).
\end{equation}
Let $Q_{\text{hist}}$ denote an orthonormal basis of the row span of $A_{\text{hist}}$ (e.g., via QR decomposition).
We project the probe onto the orthogonal complement of the historical subspace:
\begin{equation}
A_{\text{diff}}
=
A_{\text{probe}}
\big(I - Q_{\text{hist}}^{\top}Q_{\text{hist}}\big),
\label{eq:adiff}
\end{equation}
where $A_{\text{diff}}$ represents update directions required by $\mathcal{T}_k$ that are \emph{not} spanned by the accumulated skill space.
This geometric separation matches \texttt{TASER}'s goal: expand capacity only along minimally interfering directions under the current-data-only constraint.

\paragraph{Step C: Adaptive Capacity Sizing.}
Finally, we choose the smallest rank that preserves most of the residual signal, and allocate that as new skill capacity.
We perform singular value decomposition on the residual matrix:
\begin{equation}
A_{\text{diff}} = U\Sigma V^{\top}, 
\qquad
\Sigma=\mathrm{diag}(\sigma_1,\sigma_2,\dots).
\end{equation}
We choose the minimal rank that covers a target fraction of spectral energy:
\begin{equation}
r_{\text{new}}
=
\min\left\{
t \;\middle|\;
\frac{\sum_{i=1}^{t}\sigma_i^2}{\sum_{j=1}^{\text{rank}(A_{\text{diff}})}\sigma_j^2}\ge \theta_{\text{energy}}
\right\},
\label{eq:rnew}
\end{equation}
where $\theta_{\text{energy}}\in(0,1)$ is an energy coverage threshold.

We then instantiate the new capacity for task $\mathcal{T}_k$ as a single atomic skill module (parameterized via LoRA) with rank $r_{\text{new}}$. Let $\theta^{\text{new}}_k$ denote the parameters of the newly allocated atomic-skill module. The skill library is subsequently updated as $\mathcal{S}_{k} = \mathcal{S}_{k-1} \cup \{\theta^{\text{new}}_k\}$. 
After allocating the new capacity, we train only this newly added skill and the dynamic router, strictly freezing the backbone $\theta_0$ and all historical skills in $\mathcal{S}_{k-1}$ to prevent catastrophic forgetting. Specifically, we optimize $\theta^{\text{new}}_k$ and the router parameters $\phi$ (detailed in Sec.~\ref{sec:sdr}) on the current task data $\mathcal{D}_k$:
\begin{equation}
\min_{\theta^{\text{new}}_k,\,\phi}\;
\mathbb{E}_{(x,y)\sim\mathcal{D}_k}
\Big[
\mathcal{L}_k\big(f(x;\theta_0, \mathcal{S}_{k}, \phi), y\big)
\Big].
\end{equation}

Intuitively, when the new task is far from the historical skill space, $A_{\text{diff}}$ retains larger residual energy and yields a larger $r_{\text{new}}$; when tasks are related, the residual is small and the model relies more on reusing existing skills.

\subsection{Orthogonality-Enhanced Skill Detection}
\label{sec:oesd}
Under high task heterogeneity, \texttt{TASER} must preserve the \emph{atomicity} of each skill so that newly added capacity remains semantically distinct, independently reusable, and minimally interfering with previously learned skills. To this end, when training the newly instantiated LoRA skill for task $\mathcal{T}_k$, we impose explicit orthogonality constraints on its down projection matrix. Let $A_{\text{new}}$ denote the down-projection matrix of the newly allocated LoRA module and $A_{\text{hist}}$ denote the historical down-projection matrix, both defined in Sec.~\ref{sec:asil}. We optimize the following objective on the current task data $(x,y)\sim\mathcal{D}_k$:
\begin{equation}
\mathcal{L}_{\text{total}}
=
\mathcal{L}_{\text{task}}
+
\lambda_1 \mathcal{L}_{\text{inter}}
+
\lambda_2 \mathcal{L}_{\text{intra}},
\label{eq:total_loss}
\end{equation}
where $\mathcal{L}_{\text{task}}=\mathcal{L}_k\big(f(x;\theta_0,\mathcal{S}_{k},\phi),y\big)$ is the task specific loss defined by the output structure of $\mathcal{T}_k$.

\paragraph{Inter-Task Orthogonality.}
To isolate the new skill from the accumulated skill space, we penalize the correlation between $A_{\text{new}}$ and $A_{\text{hist}}$:
\begin{equation}
\mathcal{L}_{\text{inter}}
=
\left\lVert
A_{\text{new}} A_{\text{hist}}^{\top}
\right\rVert_F^2.
\label{eq:inter_ortho}
\end{equation}
Minimizing Eq.~\eqref{eq:inter_ortho} drives each row vector in $A_{\text{new}}$ to be orthogonal to the row span of $A_{\text{hist}}$, encouraging the newly allocated capacity to capture task specific directions that are not representable by existing skills. This directly supports CL under the current data only constraint by reducing representational overlap, and therefore mitigating interference with historical skills.

\paragraph{Intra-Task Orthogonality.}
When the allocated rank $r_{\text{new}}$ is greater than one, we further enforce diversity inside the new skill itself by encouraging row wise orthogonality and unit norm:
\begin{equation}
\mathcal{L}_{\text{intra}}
=
\left\lVert
A_{\text{new}} A_{\text{new}}^{\top} - I
\right\rVert_F^2,
\label{eq:intra_ortho}
\end{equation}
where $I\in\mathbb{R}^{r_{\text{new}}\times r_{\text{new}}}$ is the identity matrix. Eq.~\eqref{eq:intra_ortho} prevents rank collapse by discouraging redundant rows in $A_{\text{new}}$, improving numerical stability and ensuring that the newly introduced atomic capacity spans multiple distinct directions when needed by a highly divergent task.

During task $\mathcal{T}_k$, we freeze the backbone $\theta_0$ and all historical skills in $\mathcal{S}_{k-1}$, and update only $\theta_k^{\text{new}}$ together with the router parameters $\phi$ (Sec.~\ref{sec:sdr}) by minimizing Eq.~\eqref{eq:total_loss}. The two orthogonality terms jointly serve as an explicit skill detection mechanism: $\mathcal{L}_{\text{inter}}$ suppresses redundancy with historical skills, while $\mathcal{L}_{\text{intra}}$ enforces diversity among the newly added directions, thereby preserving skill atomicity and improving reuse across heterogeneous task streams.

\subsection{Skill Dynamic Routing}
\label{sec:sdr}
As \texttt{TASER} incrementally expands a library of atomic skills, activating all skills for every input becomes computationally inefficient and may introduce interference, especially under high task heterogeneity. We therefore propose \emph{Skill Dynamic Routing}, a lightweight gating mechanism that selects and composes a sparse subset of skills conditioned on the current input, enabling both task specific specialization and cross task reuse.

We implement each atomic skill as a LoRA-style update and keep the backbone frozen. For a given transformer layer (indices omitted for clarity), let $W_0$ denote the frozen backbone linear map, and let the up-to-date skill matrices during task $\mathcal{T}_k$ be
\begin{equation} 
A_{\text{full}} = \mathrm{Concat}(A_1,\dots,A_{M_k}), \qquad B_{\text{full}} = \mathrm{Concat}(B_1,\dots,B_{M_k}), \end{equation}
where $M_k=M_{k-1}+r_{\text{new}}$.

Given an input activation vector $x$, the router $R_\phi$ outputs a multi-skill gating vector $\boldsymbol{\alpha}(x)\in[0,1]^{M_k}$ (where $M_k = M_{k-1} + r_{\text{new}}$), and the routed layer output is computed as
\begin{equation}
h_{\text{out}}
=
W_0 x
+
B_{\text{full}}
\Big(
\boldsymbol{\alpha}(x) \odot (A_{\text{full}}x)
\Big).
\label{eq:dsr_forward}
\end{equation}
where $\odot$ is the Hadamard product. Eq.~\eqref{eq:dsr_forward} can be viewed as an input dependent selection of accumulated atomic skills, where the gating vector $\boldsymbol{\alpha}(x)$ selectively amplifies or suppresses the individual contribution of each acquired atomic skill based on the current input context. We use a Sigmoid gate rather than Softmax since multiple skills can be simultaneously helpful for a heterogeneous task.

The router is designed to be lightweight so that routing does not dominate computation. Specifically, we use a two layer MLP with a bottleneck:
\begin{equation}
\boldsymbol{\alpha}(x)
=
\sigma\Big(
W_2\,\mathrm{ReLU}(W_1 x + b_1) + b_2
\Big),
\label{eq:router}
\end{equation}
where $\phi=\{W_1,b_1,W_2,b_2\}$ are router parameters and the hidden dimension is reduced by a bottleneck ratio $r_{\text{ratio}}$ (e.g., $4$). This design supports multi skill activation while keeping the routing module parameter efficient.

To ensure interference free retrieval from the skill library, we train routing under a frozen backbone and a frozen historical skill bank. Concretely, during the joint optimization phase for task $\mathcal{T}_k$, we strictly freeze $\theta_0$ and all historical skills in $\mathcal{S}_{k-1}$. The router $\phi$ and the newly instantiated skill $\theta^{\text{new}}_k$ are updated simultaneously by minimizing the following objective:

\begin{equation}
\min_{\theta^{\text{new}}_k,\,\phi}\;
\mathbb{E}_{(x,y)\sim\mathcal{D}_k}
\Big[
\mathcal{L}_k\big(f(x;\theta_0, \mathcal{S}_{k}, \phi), y\big)
+
\lambda_{\text{sparse}}
\|\boldsymbol{\alpha}(x)\|_1
\Big],
\label{eq:router_obj}
\end{equation}

where $\mathcal{L}_k(\cdot)$ matches the output structure of task $\mathcal{T}_k$ and $\lambda_{\text{sparse}}$ controls sparsity. The $L_1$ penalty encourages the router to activate only the minimal set of relevant atomic skills for each input, improving efficiency and reducing noise from unrelated skills. This dynamic routing mechanism is essential in \texttt{TASER} for handling highly heterogeneous task streams, where different inputs within the same task may require different mixtures of reusable capabilities.

\section{Benchmark Construction}
\label{subsec:benchmark}

Existing continual learning evaluations predominantly assume similarity across sequential tasks, leaving a critical gap in assessing how models adapt to highly divergent cognitive demands. To rigorously evaluate continual learning under these high task heterogeneity conditions, we conduct experiments on our newly constructed benchmark, \texttt{HeteroCLBench}.



\subsection{Data Collection and Analysis}
\texttt{HeteroCLBench} is a skill-incremental benchmark curated primarily from Super-NaturalInstructions~\cite{wang2022super}. It is designed to operationalize \textit{high task heterogeneity} in lifelong learning, providing a stringent testbed for catastrophic forgetting and capacity allocation.

To construct a taxonomy of modern LLM cognitive capabilities, we synthesize task definitions from foundational evaluation suites (e.g., BIG-Bench, HELM). Focusing on automatically evaluable abilities, we distill this space into 9 cognitive macro-dimensions: (1) Language Understanding, (2) Mathematical \& Logical Reasoning, (3) Knowledge \& Factuality, (4) Code Generation, (5) Multilingual Capability, (6) Safety \& Ethics, (7) Interaction \& Dialog, (8) Emotional Intelligence, and (9) Planning \& Decision Making. From these, we select 19 functionally distinct tasks representing divergent input-output paradigms, jointly mapped to 24 fine-grained sub-categories. For standardized reporting, we assign semantic abbreviations to these tasks (e.g., substituting \texttt{task074} with SQuAD (QG)).

To ensure balanced evaluation without dataset scale biases, we sample a maximum of 5,000 instances for each of the 19 selected tasks. These instances are partitioned into training, validation, and test sets following an 8:1:1 ratio. During experiments, the tasks are organized into a sequential stream $ \mathcal{T} = \{T_1, \dots, T_{19}\} $. To maximize the risk of representation interference and rigorously test skill isolation, the ordering of tasks within the stream is randomized across varying cognitive modes.

\begin{table}[t!]
\centering
\caption{Comparison of \texttt{HeteroCLBench} with existing continual learning benchmarks. }
\label{tab:comparison}
\resizebox{\textwidth}{!}{
\renewcommand{\arraystretch}{1.25} 
\begin{tabular}{l c c c c c c | >{\columncolor{blue!5}\bfseries}c }
\toprule
\rowcolor{gray!10} 
\textbf{Feature} & \textbf{Standard CL} & \textbf{LNT} & \textbf{CLIF} & \textbf{TRACE} & \textbf{CITB} & \textbf{TiC-LM} & \texttt{HeteroCLBench} \\
\midrule
Task Type Annotation & \xmark & \cmark & \cmark & \cmark & \cmark & \cmark & \cmark \\
Diverse Task Formats & \xmark & \xmark & \xmark & \cmark & \cmark & \cmark & \cmark \\
Non-English Tasks & \xmark & \cmark & \xmark & \cmark & \xmark & \cmark & \cmark \\
Multiple Professional Domains & \xmark & \cmark & \xmark & \cmark & \cmark & \cmark & \cmark \\
\midrule
Number of Tasks & 5 & 15 & 6 & 8 & 13 & 4 & 19 \\
Macro-dimensions & 2 & 3 & 4 & 3 & 4 & 3 & 9 \\
Sub-categories & 3 & 5 & 7 & 6 & 7 & 6 &  24\\
\bottomrule
\end{tabular}
}
\end{table}

\subsection{Comparison with Existing Benchmarks}
As shown in Table~\ref{tab:comparison}, \texttt{HeteroCLBench} stands out from existing continual learning benchmarks in three aspects.
First, it achieves broad functional coverage where prior benchmarks involve structural compromises. For example, although Large Number of Tasks (LNT) \cite{wang2023orthogonal_olora_LNT} includes a long sequence of 15 tasks, it focuses only on text classification. Other CL benchmarks, such as Standard CL \cite{wang2023orthogonal_olora_LNT,zhang2015character_standcl} and CLIF \cite{jin2021learn_clif}, also miss important settings, including non-English evaluation and professional multi-domain scenarios. In contrast, \texttt{HeteroCLBench} covers all of these key aspects.
Second, our benchmark introduces unprecedented cognitive heterogeneity. Existing benchmarks, including TRACE \cite{wang2023trace}, CITB \cite{zhang2023citb}, and TiC-LM \cite{li2025tic}, evaluate only 2 to 4 macro-dimensions. By comparison, \texttt{HeteroCLBench} covers 9 macro-dimensions, enabling evaluation across more functionally distinct cognitive abilities rather than simple distribution shifts.
Third, \texttt{HeteroCLBench} provides an exceptionally fine-grained evaluation framework. It includes 19 functionally distinct tasks and 24 sub-categories, far exceeding the previous maximum of 7 sub-categories reported in CLIF and CITB. This greater granularity increases semantic diversity within the task stream and more rigorously tests model capacity allocation and forward transfer.

\section{Experimental Setups}

\subsection{Baselines}
To comprehensively evaluate \texttt{TASER}, we compare it against a wide range of continual learning methods and bounds: \textbf{Sequential Fine-tuning (Seq FT)} naively fine-tunes the model sequentially without constraints; \textbf{EWC}~\cite{huszar2018note_ewc} mitigates forgetting by penalizing changes to parameters critical for previous tasks; \textbf{Experience Replay}~\cite{rolnick2019experience_replay} maintains a small episodic memory buffer of past examples for rehearsal; \textbf{LAMOL}~\cite{sun2019lamol} uses generative replay to synthesize historical data without an external buffer; \textbf{I-LoRA}~\cite{li2025analyzing} reduces forgetting by linearly interpolating between old and new LoRA weights; \textbf{MoE-LoRA}~\cite{chen2024coin} employs a Mixture-of-Experts architecture to dynamically allocate LoRA experts; and \textbf{CL-MoE}~\cite{huai2025cl} enhances MoE stability through dual momentum routing. Finally, \textbf{Multitask Learning (MTL)} trains a single model on all tasks simultaneously, serving as a commonly used empirical upper-bound reference for CL performance~\cite{huai2025cl,shi2025continual_survey}.

\begin{table}[t!]
\centering
\caption{The main results over \texttt{HeteroCLBench} in terms of AP and BWT.}
\label{tab:result_llama3_cross}
\resizebox{\textwidth}{!}{
\renewcommand{\arraystretch}{1.15} 
\begin{tabular}{l | ccccccc | >{\columncolor{blue!5}}c | c}
\toprule
\multirow{2}{*}{\textbf{Task}} & \multicolumn{7}{c|}{\textbf{Continual Learning Baselines}} & \multirow{1}{*}{Ours} & \textbf{Upper Bound} \\
\cmidrule{2-10}
& \textbf{Seq FT} & \textbf{EWC} & \textbf{Replay} & \textbf{LAMOL} & \textbf{I-LoRA} & \textbf{MoE-LoRA} & \textbf{CL-MoE} & \textbf{TASER} & \textbf{MTL} \\
\midrule
SQuAD (QG)   & 16.49  & 17.76  & 36.69  & 16.50  & 37.20  & 13.38  & 15.27  & \textbf{40.11}  & 38.38 \\
XSum         & 25.79  & 26.44  & 30.64  & 23.87  & 31.32  & 24.99  & 25.14  & \textbf{33.86}  & 33.29 \\
CoNLL-02     & 46.78  & 46.78  & \textbf{89.48}  & 45.28  & 83.05  & 46.57  & 46.78  & 85.41  & 93.35 \\
MathQA       & 31.80  & 34.60  & 38.80  & 31.80  & \textbf{40.80}  & 33.00  & 33.40  & 40.00  & 43.20 \\
DM Math      & 63.60  & 67.00  & \textbf{71.00}  & 66.60  & 69.00  & 67.00  & 68.20  & \textbf{71.00}  & 84.80 \\
BoolQ        & 70.67  & 80.33  & \textbf{85.44}  & 76.97  & 84.59  & 78.38  & 76.45  & 81.40  & 87.25 \\
CREAK        & 87.80  & 86.40  & \textbf{95.80}  & 91.20  & 95.00  & 92.20  & 88.00  & 95.00  & 96.20 \\
CodeComp     & 29.40  & 37.80  & \textbf{72.00}  & 42.80  & 62.60  & 44.60  & 45.00  & 64.00  & 68.80 \\ 
Code2Text    & 26.68  & 22.91  & 35.30  & 25.42  & \textbf{36.29}  & 29.58  & 28.35  & 36.12  & 37.13 \\
TED (fa-he)  & 49.01  & 50.17  & 53.01  & 45.54  & 54.79  & 50.93  & 50.16  & \textbf{56.42}  & 56.06 \\
ALT (ja-th)  & 41.73  & 43.76  & 47.36  & 44.84  & 48.53  & 44.66  & 45.72  & \textbf{51.08}  & 50.33 \\
Scruples     & 73.40  & 72.20  & 71.80  & 66.40  & \textbf{75.20}  & 70.00  & 72.40  & 73.20  & 76.40 \\
Jigsaw       & 80.20  & 86.20  & 96.60  & 81.40  & 95.80  & 89.60  & 89.60  & \textbf{100.00} & 97.60 \\
MuTual       & 71.80  & 75.20  & 82.20  & 80.80  & 80.20  & 78.80  & 77.20  & \textbf{83.80}  & 87.80 \\
AirDialog    & 42.86  & 54.80  & 58.27  & 47.02  & 57.68  & 51.37  & 46.99  & \textbf{61.37}  & 60.71 \\
TwitterEmo   & 55.98  & 64.14  & \textbf{82.51}  & 82.22  & 81.34  & 76.09  & 74.05  & 80.76  & 86.88 \\
StoryCS      & 33.14  & 33.70  & 31.37  & 33.28  & 30.39  & 35.87  & \textbf{36.32}  & 35.88  & 33.23 \\
RecipeNLG    & 36.00  & 36.35  & 35.81  & 37.56  & 35.83  & \textbf{38.84}  & 38.69  & 36.58  & 37.20 \\
HellaSwag    & 81.40  & 81.20  & 76.40  & 80.20  & 75.00  & \textbf{82.40}  & 81.40  & 77.00  & 79.60 \\
\midrule
\rowcolor{gray!10}
\textbf{AP}  & 50.76  & 53.56  & 62.66  & 53.67  & 61.82  & 55.17  & 54.69  & \textbf{63.31}  & 65.70 \\
\rowcolor{gray!10}
\textbf{BWT} & -13.54 & -13.13 & -2.03  & -14.45 & -1.56  & -11.47 & -15.13 & \textbf{-0.94}  & -     \\
\bottomrule
\end{tabular}
}
\end{table}

\subsection{Implementation Details}
We use {Llama-3-8B-Instruct}~\cite{grattafiori2024llama} as the backbone model and implement all methods in {PyTorch}, with experiments conducted on {NVIDIA RTX 4090} GPUs. Unless otherwise specified, we optimize the trainable parameters using {AdamW} with a batch size of 32 and a learning rate of $1\times10^{-4}$. For parameter-efficient fine-tuning baselines, the LoRA rank is set to $r=8$ with scaling factor $\alpha=16$. Baseline-specific settings are as follows: {EWC} uses $\lambda=10$ and 200 samples per task to estimate the Fisher Information Matrix; Experience Replay adopts a replay ratio of 0.3; {LAMOL} generates 200 pseudo-samples for each previous task; {I-LoRA} uses an EMA momentum of 0.25 and a consistency weight of 1.0; and {MoE-LoRA} and {CL-MoE} use 4 experts, with {CL-MoE} employing top-$k$ routing with $k=2$. For \texttt{TASER}, we set the probe rank budget to $r_{\text{probe}} = r_{\text{max}} = 8$ unless otherwise specified, and dynamically determine the number of new atomic skills using an energy threshold of $\theta_{\text{energy}}=0.9$.


\subsection{Evaluation Metrics}
Following \cite{lopez2017gradient_gem}, we report Average Performance (AP) as the overall performance metric and Backward Transfer (BWT) to measure how learning new tasks affects the performance on previously learned tasks. Let $R_{i,j}$ denote the performance of the model evaluated on task $j$ after sequentially learning up to task $i$. After training on a sequence of $N$ tasks, AP and BWT are formally defined as:
\begin{equation}
\text{AP} = \frac{1}{N} \sum_{j=1}^{N} R_{N,j}, \qquad \text{BWT} = \frac{1}{N-1} \sum_{j=1}^{N-1} \big(R_{N,j} - R_{j,j}\big).
\end{equation}
To calculate $R_{i,j}$ for classification and multiple-choice tasks, we use Exact Match (Accuracy); for open-ended generation tasks, we use ROUGE-L, following Natural-Instructions. To make scores comparable across the highly heterogeneous \texttt{HeteroCLBench}, all task metrics are normalized to the range of $[0, 100]$ before computing the final AP and BWT.

\section{Experimental Analysis}
\subsection{Main Results}
\label{subsec:main_results}
As shown in Table~\ref{tab:result_llama3_cross}, \texttt{TASER} achieves the superior Average Performance (AP) of 63.31 on \texttt{HeteroCLBench}, outperforming all continual learning baselines and approaching the multi-task learning (MTL) upper bound of 65.70. It also delivers strong results across diverse task categories, with notable gains on tasks such as \texttt{ALT (ja-th)} and \texttt{MuTual}, which highlights its robust plasticity and capacity to adapt to highly heterogeneous task streams.
Furthermore, \texttt{TASER} effectively addresses catastrophic forgetting, achieving the best Backward Transfer (BWT) of -0.94. This represents a significant improvement over sequential fine-tuning (-13.54) and other modular baselines like \texttt{CL-MoE} (-15.13). These results demonstrate that the proposed atomic skill expansion and orthogonality-enhanced detection mechanisms successfully maintain stability by ensuring skills remain semantically distinct and reusable.

\begin{table}[tbp]
    \centering
    \caption{Ablation study of the core modules in \texttt{TASER}. $\Delta$ indicates the performance variation compared to the full framework.}
    \label{tab:ablation}
    \renewcommand{\arraystretch}{1.15} 
    \begin{tabular}{lcccc}
        \toprule
        \textbf{Model Variant} & \textbf{AP ($\uparrow$)} & \textbf{$\Delta_{\text{AP}}$} & \textbf{BWT ($\uparrow$)} & \textbf{$\Delta_{\text{BWT}}$} \\
        \midrule
        \textbf{Full \texttt{TASER} framework} & \textbf{63.31} & - & \textbf{-0.94} & - \\
        \midrule
        \quad w/o Atomic Skill Incremental Learning & 57.14 & -6.17 & -2.69 & -1.75 \\
        \quad w/o Orthogonality-enhanced Skill Detection & 62.74 & -0.57 & -1.4 & -0.46 \\
        \quad w/o Skill Dynamic Routing & 44.82 & -18.49 & -5.39 & -4.45 \\
        \bottomrule
    \end{tabular}
\end{table}

\subsection{Ablation Studies}
To investigate the contribution of each core component in \texttt{TASER}, we conduct an ablation study as summarized in Table~\ref{tab:ablation}. The results demonstrate that all three modules are critical, as the removal of any single component leads to a consistent decline in both the composite AP and BWT metrics. Specifically, removing the Skill Dynamic Routing module causes the most substantial performance collapse, resulting in an absolute drop of 18.49 in AP and 4.45 in BWT. This highlights its essential role in managing task interference through precise skill activation. The absence of Atomic Skill Incremental Learning leads to an absolute reduction of 6.17 in AP, confirming that adaptive capacity expansion is vital for capturing the diverse reasoning patterns inherent in heterogeneous task streams. Finally, the Orthogonality-enhanced Skill Detection module is shown to be necessary for maintaining representational purity, as its removal incurs an absolute decline of 0.57 in AP and 0.46 in BWT.

\begin{figure}[t!]
    \centering
    \includegraphics[width=\textwidth]{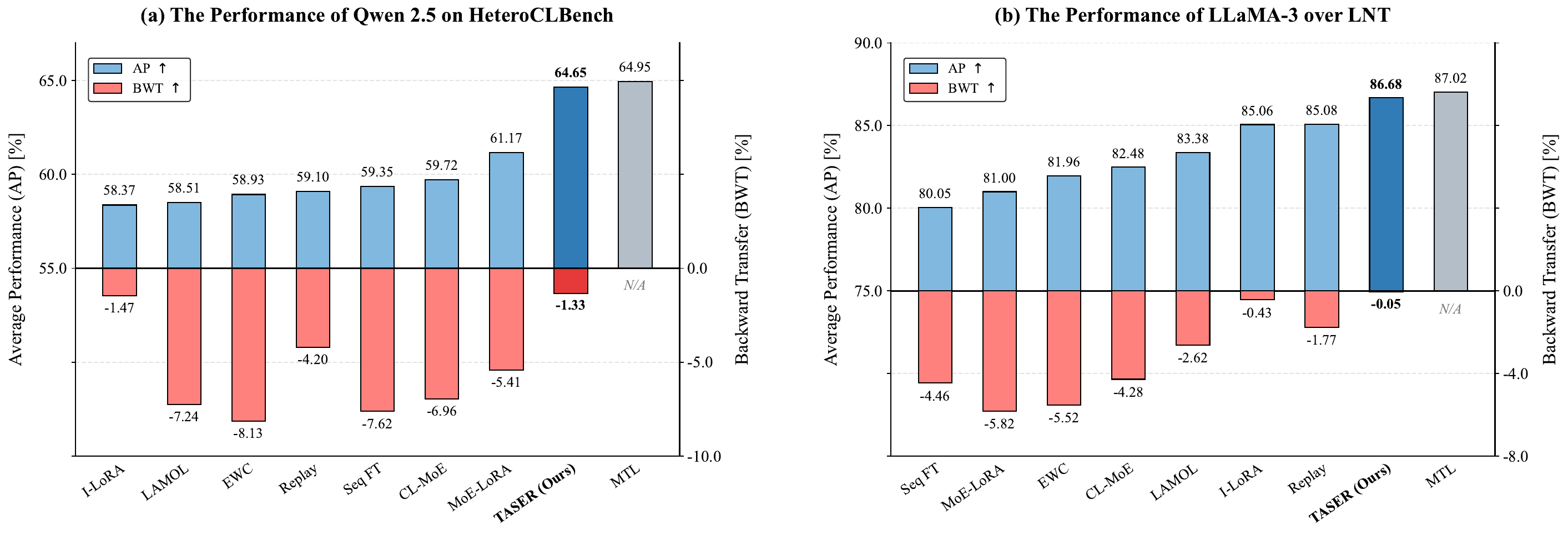}
    \caption{The performance of robustness analysis. \textbf{(a)} Evaluation of Qwen-2.5 on \texttt{HeteroCLBench}. \textbf{(b)} Evaluation of LLaMA-3 on LNT. The upward bars (blue) represent the AP of each method, while the downward bars (red) represent the BWT.}
    \label{fig:main_results}
\end{figure}

\begin{figure}[t!]
    \centering
    \includegraphics[width=\textwidth]{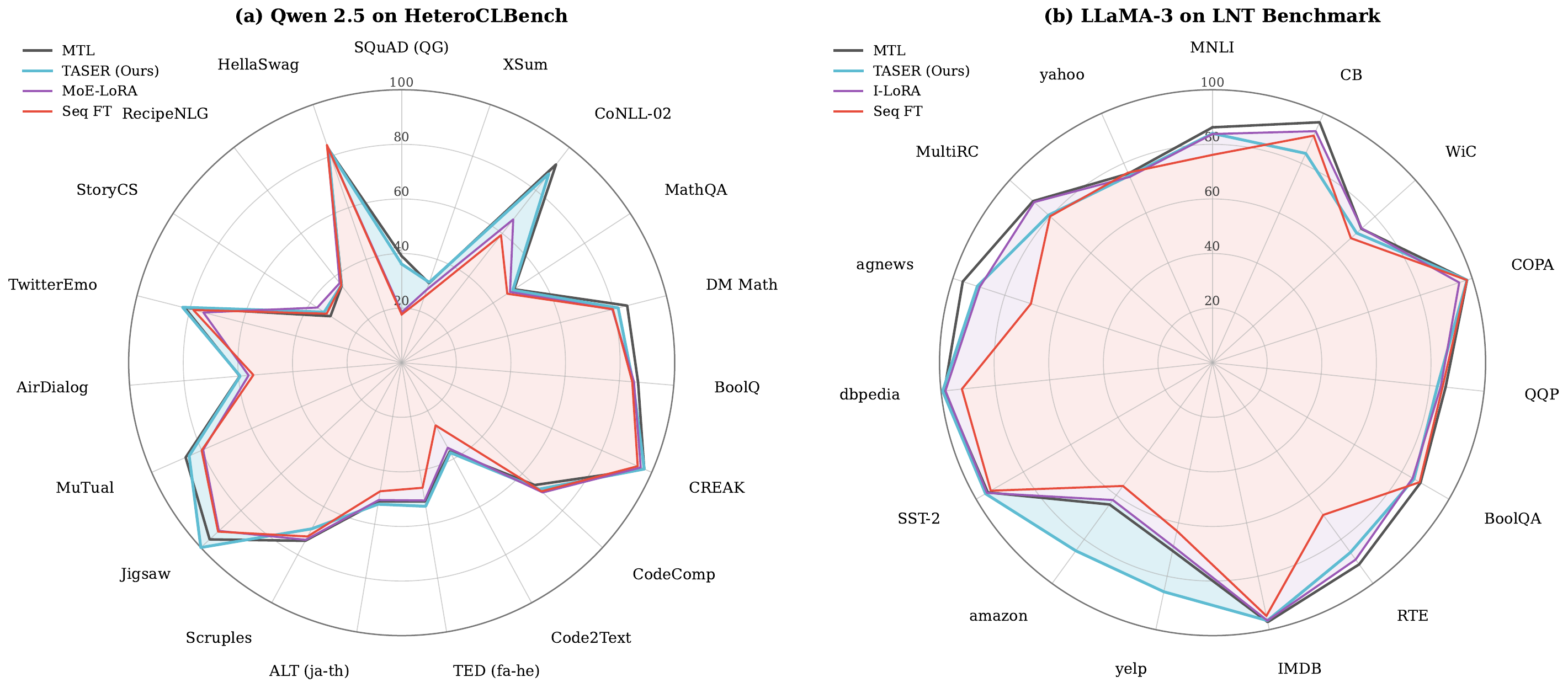}
    \caption{Task-level performance across diverse cognitive dimensions. \textbf{(a)} Qwen-2.5 on \texttt{HeteroCLBench}. \textbf{(b)} LLaMA-3 on LNT. Alongside our \texttt{TASER} and the \texttt{Seq FT} baseline, we plot \texttt{MTL} and the strongest CL baseline (\texttt{MoE-LoRA} and \texttt{I-LoRA}, respectively).}
    \label{fig:radar_comparison}
\end{figure}

\subsection{Robustness Analysis}

To validate the robustness of \texttt{TASER} across different model architectures, we evaluate its performance using Qwen-2.5~\cite{yang2024qwen2technicalreport} on \texttt{HeteroCLBench} (see Fig.~\ref{fig:main_results} and Fig.~\ref{fig:radar_comparison}(a)). \texttt{TASER} achieves an AP of 64.65, closely approaching the Multi-Task Learning (MTL) upper bound of 64.95, and outperforming the strongest baseline, \texttt{MoE-LoRA} (61.17). Although \texttt{Seq FT} surpasses several CL baselines on Qwen-2.5, its BWT remains much worse (-7.62), indicating poor stability under continual updates. This likely reflects strong instruction-style priors already present in the backbone. In contrast, \texttt{TASER} preserves such priors while maintaining substantially better stability, achieving the best BWT of -1.33.

Extending our evaluation to assess cross-benchmark generalization, we test \texttt{TASER} on the LNT task sequence~\cite{wang2023orthogonal_olora_LNT} using LLaMA-3 (Fig.~\ref{fig:main_results} and Fig.~\ref{fig:radar_comparison}(b)). Under this setting, \texttt{TASER} reaches an AP of 86.68 alongside an almost negligible BWT of -0.05, nearly matching the MTL joint-training performance of 87.02. As illustrated by the radar plots, \texttt{TASER} maintains a highly balanced performance profile across all evaluated cognitive dimensions compared to competing baselines. Collectively, these results confirm that the architectural advantages of \texttt{TASER} generalize robustly across diverse base models and continuous learning environments.

\begin{figure}[t!]
    \centering
    \includegraphics[width=\textwidth]{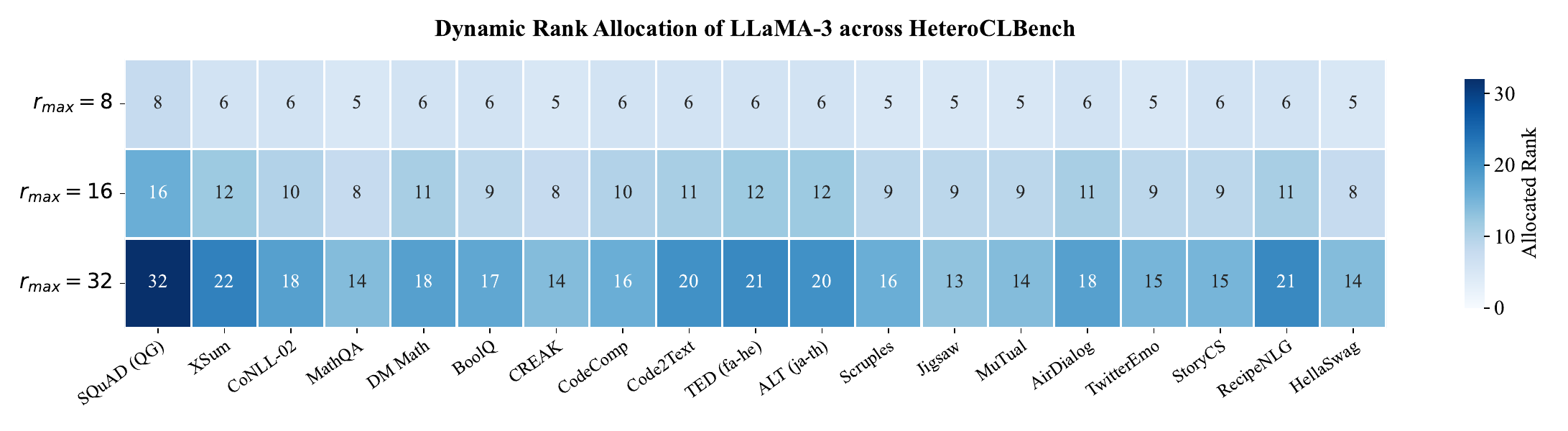}
    \caption{Visualization of the dynamic rank allocation for LLaMA-3 over \texttt{HeteroCLBench}. 
    }
    \label{fig:dynamic_rank_heatmap}
\end{figure}

\subsection{Further Analysis}
\textbf{Analysis of Dynamic Rank Allocation.}
As illustrated in Figure \ref{fig:dynamic_rank_heatmap}, the dynamic rank allocation of \texttt{TASER} exhibits remarkable scale-invariant consistency across different capacity budgets ($r_{max} \in \{8, 16, 32\}$), validating the robustness of our SVD-based routing as a stable estimator of task complexity. The allocation reflects the intrinsic representational demands of the tasks: generative and structurally demanding tasks, such as \texttt{SQuAD (QG)} and cross-lingual translation (\texttt{TED}, \texttt{ALT}), consistently receive higher ranks to accommodate significant distribution shifts. Conversely, reasoning or classification-oriented tasks like \texttt{MathQA} and \texttt{Jigsaw} consume minimal capacity by effectively reusing existing semantic priors. This adaptive behavior provides strong empirical evidence that \texttt{TASER} successfully prevents capacity redundancy while ensuring sufficient plasticity for highly heterogeneous task streams.

\begin{figure}[t!]
    \centering
    \includegraphics[width=0.85\textwidth]{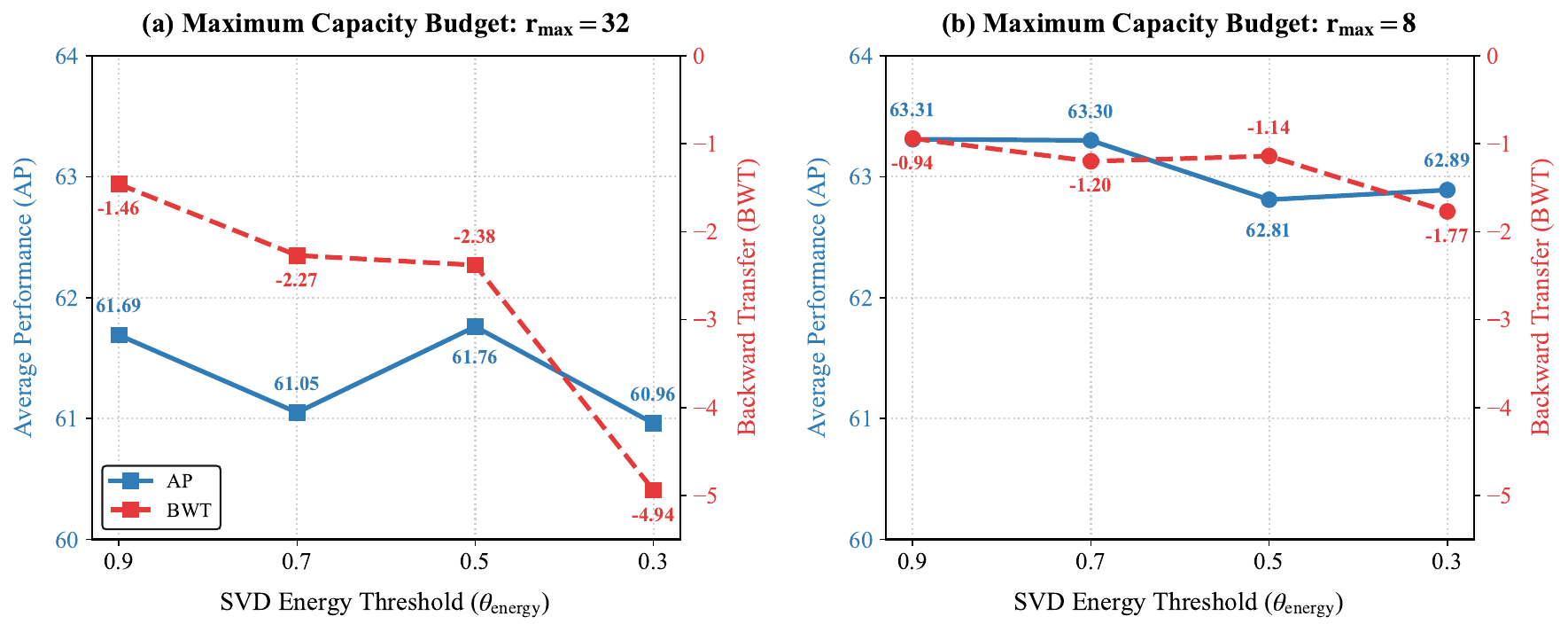}
    \caption{Sensitivity analysis of the SVD energy threshold ($\theta_{\text{energy}}$). The chart illustrates the impact of decreasing the energy threshold on both AP (left axis in blue) and BWT (right axis in red) under different capacity budgets.}
    \label{fig:sensitivity_svd}
\end{figure}

\textbf{Sensitivity Analysis of Hyper-Parameter.}
To investigate the robustness of \texttt{TASER} against varying hyperparameters and to explicitly verify whether its performance gains stem merely from capacity inflation, we conduct comprehensive sensitivity analyses.
We first evaluate the impact of the SVD energy threshold ($\theta_{\text{energy}}$) under different initial probe rank capacities ($r_{\text{max}} \in \{8, 32\}$). As shown in Fig.~\ref{fig:sensitivity_svd}, \texttt{TASER} demonstrates remarkable stability when configured with a compact probe rank ($r_{\text{max}}=8$), maintaining both high AP and near-zero BWT across different thresholds. Conversely, an oversized initial budget ($r_{\text{max}}=32$) leads to severe performance degradation and catastrophic forgetting (indicated by the sharp drop in BWT) when $\theta_{\text{energy}}$ decreases. This indicates that overly permissive spectral thresholds in an oversized subspace retain redundant update directions, which inevitably interfere with previously learned skills.


\section{Conclusion}
We addressed the challenge of continual learning for LLMs under high task heterogeneity by proposing \texttt{TASER}, a modular framework integrating dynamic atomic skill expansion, orthogonality-enhanced skill detection, and dynamic routing. To rigorously evaluate this setting, we introduced \texttt{HeteroCLBench}, comprising 19 functionally diverse tasks across 9 cognitive dimensions. Experiments demonstrate that \texttt{TASER} effectively balances plasticity and stability, significantly reducing catastrophic forgetting and approaching the multi-task learning upper bound. By dynamically allocating capacity based on task divergence and ensuring skill independence, \texttt{TASER} offers a robust and scalable solution for adapting LLMs to complex, real-world lifelong learning scenarios.

\section*{Acknowledgments}
The authors wish to thank the reviewers for their helpful comments and suggestions.
This research is funded by the National Key Research and Development Program of China Grant (No. 2024YFC3308500), the National Natural Science Foundation of China (No. 62477010, No.62577022 and No.62307028), Shanghai Science and Technology Innovation Action Plan (No. 24YF2710100), and the opening funding of the State Key Laboratory of Disaster Reduction in Civil Engineering (Grant No. SLDRCE24-03).

%
%
\bibliographystyle{splncs04}
\bibliography{main}
\end{document}


\title{Supplementary Material for: Task-Differentiated Atomic Skill Expansion and Routing for Continual Learning Across Highly Heterogeneous Tasks}
\titlerunning{Supplementary Material for TASER}
\author{} 
\institute{}
\maketitle
\appendix

\renewcommand{\arraystretch}{1.1}

\begin{table}[htbp]
\centering
\caption{The 9 macro-dimensions of NLP evaluation within \texttt{HeteroCLBench}, detailing the subcategories, definitions, and typical tasks. Sources are indicated in brackets.}
\label{tab:macro_dimensions}
\scriptsize
\setlength{\tabcolsep}{3pt}
\begin{tabular}{>{\raggedright\arraybackslash}p{2.2cm} >{\raggedright\arraybackslash}p{3.2cm} >{\raggedright\arraybackslash}p{3.8cm} >{\raggedright\arraybackslash}p{2.6cm}}
\toprule
\textbf{Dimension} & \textbf{Subcategory} & \textbf{Brief Definition} & \textbf{Typical Tasks [Sources]} \\
\midrule
\textbf{Language Understanding \& Generation} 
& Text completion, QA, summarization, NER, text rewriting 
& Understand input language to generate natural output or extract key information. 
& QA (SQuAD), Summarization \newline [FLAN, SuperNI] \\
\addlinespace

\textbf{Mathematical Logic} 
& Deductive/inductive reasoning, axiomatic proof, symbolic computation 
& Perform logical deduction, mathematical proof, or problem-solving based on formal rules. 
& MathQA, GSM8K, ProofWriter \newline [BIG-Bench] \\
\addlinespace

\textbf{Knowledge \& Factuality} 
& General knowledge recall, factuality, evidence-based inference 
& Perform inference and retrieval based on world knowledge and facts. 
& TriviaQA, FEVER \newline [HELM, BIG-Bench] \\
\addlinespace

\textbf{Code Gen. \& Understanding} 
& Code completion, code translation, code debugging 
& Understand natural language instructions to generate or modify programming code. 
& HumanEval, MBPP \newline [BIG-Bench] \\
\addlinespace

\textbf{Multilingual Capability} 
& Cross-lingual understanding and generation 
& Understand, translate, or generate text across multiple languages. 
& XNLI, WMT Tasks \newline [HELM, FLAN] \\
\addlinespace

\textbf{Safety \& Ethics} 
& Bias detection, toxic content detection, fairness 
& Detect and mitigate bias, inappropriate content, and unfair outputs. 
& RealToxicityPrompts, HolisticBias \newline [HELM] \\
\addlinespace

\textbf{Interaction \& Dialogue} 
& Multi-turn dialogue, long-context retention 
& Understand dialogue history and generate contextually consistent responses. 
& DSTC8, Chatbots \newline [HELM, SuperNI] \\
\addlinespace

\textbf{Emotional Intelligence} 
& Understanding\slash generating emotional content, empathy 
& Comprehend emotions and generate emotionally resonant replies. 
& EmpatheticDialogues, GoEmotions \\
\addlinespace

\textbf{Planning \& Decision Making} 
& Strategic planning, decision making in complex scenarios 
& Formulate actionable plans in multi-step or complex situations. 
& ALFWorld, ScienceWorld \\
\bottomrule
\end{tabular}
\end{table}

\begin{table}[htbp]
\centering
\caption{Detailed statistics, metric evaluation, and descriptions of the 19 diverse tasks comprising the \texttt{HeteroCLBench}.}
\label{tab:heteroclbench_details}
\scriptsize
\setlength{\tabcolsep}{3pt} 
\begin{tabular}{>{\raggedright\arraybackslash}p{1.8cm} >{\raggedright\arraybackslash}p{1.5cm} >{\raggedright\arraybackslash}p{4.0cm} >{\raggedright\arraybackslash}p{3.8cm}}
\toprule
\textbf{Task} & \textbf{Metric} & \textbf{Original Source / Task ID} & \textbf{Task Description} \\
\midrule

\multicolumn{4}{l}{\textbf{Language Understanding \& Generation}} \\
\midrule
SQuAD (QG) & ROUGE-L & \texttt{task074\_squad1.1\_} \newline \texttt{question\_generation} & Generates context-based questions. \\
XSum & ROUGE-L & \texttt{task1290\_xsum\_} \newline \texttt{summarization} & Generates concise news summaries. \\
CoNLL-02 & Accuracy & \texttt{task1544\_conll2002\_ner} & Named entity recognition. \\
\midrule

\multicolumn{4}{l}{\textbf{Mathematical Logic}} \\
\midrule
MathQA & Accuracy & \texttt{task1420\_mathqa\_general} & Solves math word problems (MCQ). \\
DM Math & Accuracy & \texttt{deepmind\_mathematics\_} \newline \texttt{dataset} & Solves diverse mathematical queries. \\
\midrule

\multicolumn{4}{l}{\textbf{Knowledge \& Factuality}} \\
\midrule
BoolQ & ROUGE-L & \texttt{BoolQA} & Determines truth of factual propositions. \\
CREAK & Accuracy & \texttt{task403\_creak\_} \newline \texttt{commonsense\_inference} & Evaluates commonsense claims. \\
\midrule

\multicolumn{4}{l}{\textbf{Code Generation \& Understanding}} \\
\midrule
CodeComp & ROUGE-L & \texttt{CodeXGLUE} \newline \texttt{(CodeCompletion-token)} & Predicts next token in code execution. \\
Code2Text & ROUGE-L & \texttt{CodeXGLUE (code-to-text)} & Generates natural language for Java code. \\
\midrule

\multicolumn{4}{l}{\textbf{Multilingual Capability}} \\
\midrule
TED (fa-he) & ROUGE-L & \texttt{task1269\_ted\_} \newline \texttt{translation\_fa\_he} & Translates sentences from Persian to Hebrew. \\
ALT (ja-th) & ROUGE-L & \texttt{task1127\_alt\_ja\_th\_} \newline \texttt{translation} & Translates sentences from Japanese to Thai. \\
\midrule

\multicolumn{4}{l}{\textbf{Safety \& Ethics}} \\
\midrule
Scruples & Accuracy & \texttt{task106\_scruples\_} \newline \texttt{ethical\_judgment} & Judges which action is less ethical. \\
Jigsaw & Accuracy & \texttt{task322\_jigsaw\_} \newline \texttt{classification\_threat} & Classifies sentences for threat potential. \\
\midrule

\multicolumn{4}{l}{\textbf{Interaction \& Dialogue}} \\
\midrule
MuTual & Accuracy & \texttt{task611\_mutual\_multi\_} \newline \texttt{turn\_dialogue} & Selects the best dialogue continuation. \\
AirDialog & ROUGE-L & \texttt{task574\_air\_dialogue\_} \newline \texttt{sentence\_generation} & Fills in gaps in flight booking dialogues. \\
\midrule

\multicolumn{4}{l}{\textbf{Emotional Intelligence}} \\
\midrule
TwitterEmo & Accuracy & \texttt{task512\_twitter\_} \newline \texttt{emotion\_classification} & Classifies tweets into emotion categories. \\
StoryCS & ROUGE-L & \texttt{task293\_} \newline \texttt{storycommonsense...} & Generates emotional reactions to stories. \\
\midrule

\multicolumn{4}{l}{\textbf{Planning \& Decision Making}} \\
\midrule
RecipeNLG & ROUGE-L & \texttt{RecipeNLG} & Generates cooking instructions. \\
HellaSwag & Accuracy & \texttt{hellaswag} & Chooses the most logical situation continuation. \\
\bottomrule
\end{tabular}
\end{table}

\begin{table}[t!]
\centering
\caption{The main results of Qwen2.5 over \texttt{HeteroCLBench}.}
\label{tab:result_qwen_cross}
\resizebox{\textwidth}{!}{
\renewcommand{\arraystretch}{1.15}
\begin{tabular}{l | ccccccc | >{\columncolor{blue!5}}c | c}
\toprule
\multirow{2}{*}{\textbf{Task}} & \multicolumn{7}{c|}{\textbf{Continual Learning Baselines}} & \multirow{1}{*}{\textbf{Ours}} & \textbf{Upper Bound} \\
\cmidrule{2-10}
& \textbf{Seq FT} & \textbf{EWC} & \textbf{Replay} & \textbf{LAMOL} & \textbf{I-LoRA} & \textbf{MoE-LoRA} & \textbf{CL-MoE} & \textbf{TASER} & \textbf{MTL} \\
\midrule
SQuAD (QG)   & 17.68 & 17.53 & 35.07 & 19.23 & 34.04 & 18.43 & 28.81 & \textbf{36.16} & 39.08 \\
XSum         & 25.49 & 26.48 & 27.07 & 25.43 & 28.58 & 28.29 & 25.05 & \textbf{31.13} & 30.82 \\
CoNLL-02     & 59.22 & 58.15 & 80.68 & 54.07 & 72.74 & 66.52 & 49.78 & \textbf{87.77} & 92.06 \\
MathQA       & 46.20 & 45.80 & 40.80 & 43.40 & 41.80 & 47.40 & 45.20 & \textbf{48.60} & 49.20 \\
DM Math      & 79.80 & 80.60 & 76.20 & 81.20 & 66.40 & 79.60 & 79.80 & \textbf{81.80} & 85.20 \\
BoolQ        & 84.74 & 84.55 & 82.99 & 85.10 & 84.89 & 85.38 & \textbf{85.41} & 85.00 & 86.88 \\
CREAK        & 94.40 & 91.80 & 96.60 & 95.20 & 96.20 & 95.60 & 92.60 & \textbf{97.00} & 97.00 \\
CodeComp     & 69.60 & 68.40 & 63.20 & 62.40 & 40.00 & \textbf{70.20} & 67.40 & 68.40 & 66.20 \\
Code2Text    & 26.11 & 25.06 & 30.99 & 27.04 & 32.09 & 35.45  & 31.43 & \textbf{37.48} & 37.02 \\
TED (fa-he)  & 46.46 & 45.95 & 45.00 & 44.08 & 47.95 & 51.19 & 47.05 & \textbf{53.34} & 51.55 \\
ALT (ja-th)  & 47.76 & 47.08 & 43.67 & 47.06 & 49.26 & 51.02 & 49.17 & \textbf{52.59} & 51.51 \\
Scruples     & 72.40 & 72.60 & 71.20 & 69.60 & 72.00 & \textbf{73.80} & 70.60 & 69.20 & 74.20 \\
Jigsaw       & 91.40 & 92.80 & 90.40 & 91.00 & 92.00 & 91.00 & 90.40 & \textbf{100.00}& 95.60 \\
MuTual       & 80.00 & 75.60 & 78.80 & 80.80 & 80.00 & 79.40 & 80.60 & \textbf{85.00} & 86.40 \\
AirDialog    & 54.50 & 56.81 & 55.16 & 56.23 & 55.34 & 56.26 & 59.05 & \textbf{59.37} & 59.52 \\
TwitterEmo   & 78.71 & 77.55 & 61.81 & 76.96 & 74.05 & 74.93 & 76.09 & \textbf{82.80} & 81.92 \\
StoryCS      & 32.82 & 33.08 & 30.14 & 32.73 & 29.00 & 36.92 & \textbf{37.02} & 33.90 & 31.16 \\
RecipeNLG    & 35.98 & 35.77 & 35.39 & 36.24 & 32.31 & \textbf{36.97} & 36.82 & 35.96 & 35.74 \\
HellaSwag    & \textbf{84.40} & 84.00 & 77.80 & 83.80 & 80.40 & 83.80 & 82.40 & 82.80 & 83.00 \\
\midrule
\rowcolor{gray!10}
\textbf{AP}  & 59.35 & 58.93 & 59.10 & 58.51 & 58.37 & 61.17 & 59.72 & \textbf{64.65} & 64.95 \\
\rowcolor{gray!10}
\textbf{BWT} & -7.62 & -8.13 & -4.20 & -7.24 & -1.47 & -5.41 & -6.96 & \textbf{-1.33} & -     \\
\bottomrule
\end{tabular}
}
\end{table}

\begin{table}[htbp]
\centering
\caption{Continual learning performance of \textbf{Llama-3-8B-Instruct} on the \textbf{LNT Benchmark}. All values represent Accuracy.}
\label{tab:result_llama3_lnt}
\resizebox{\textwidth}{!}{
\renewcommand{\arraystretch}{1.15} 
\begin{tabular}{l | ccccccc | >{\columncolor{blue!5}}c | c}
\toprule
\multirow{2}{*}{\textbf{Task}} & \multicolumn{7}{c|}{\textbf{Continual Learning Baselines}} & \multirow{1}{*}{Ours} & \textbf{Upper Bound} \\
\cmidrule{2-10}
& \textbf{Seq FT} & \textbf{EWC} & \textbf{Replay} & \textbf{LAMOL} & \textbf{I-LoRA} & \textbf{MoE-LoRA} & \textbf{CL-MoE} & \textbf{TASER} & \textbf{MTL} \\
\midrule
MNLI         & 76.16 & 72.96 & \textbf{84.23} & 78.88 & 83.79 & 74.76 & 74.58 & 84.00 & 86.24 \\
CB           & 91.07 & 80.36 & 87.50 & 83.92 & \textbf{92.86} & 62.50 & 69.64 & 83.93 & 96.43 \\
WiC          & 68.18 & 65.05 & 71.40 & 68.65 & \textbf{73.51} & 59.87 & 68.34 & 71.00 & 73.35 \\
COPA         & \textbf{98.00} & 96.00 & 94.00 & 94.00 & 95.00 & 94.00 & 97.00 & \textbf{98.00} & 98.00 \\
QQP          & \textbf{84.72} & 79.28 & 83.85 & 80.55 & 84.04 & 81.22 & 82.18 & 82.80 & 85.82 \\
BoolQA       & \textbf{87.58} & 78.13 & 86.26 & 83.06 & 84.62 & 81.25 & 81.53 & 85.20 & 87.95 \\
RTE          & 68.95 & 82.31 & \textbf{89.89} & 81.23 & 89.17 & 88.81 & \textbf{89.89} & 85.92 & 91.34 \\
IMDB         & 94.86 & 96.00 & 96.69 & \textbf{96.78} & 96.62 & 96.29 & 96.13 & 96.60 & 97.16 \\
yelp         & 63.00 & 66.82 & 68.00 & 68.35 & 68.88 & 67.14 & 68.04 & \textbf{85.80} & 70.46 \\
amazon       & 55.83 & 61.37 & 62.12 & 64.32 & 62.19 & 63.05 & 64.51 & \textbf{85.20} & 64.21 \\
SST-2        & 93.81 & 93.92 & 95.07 & 94.50 & 95.41 & 94.84 & 94.50 & \textbf{96.00} & 95.07 \\
dbpedia      & 92.32 & 98.32 & 98.38 & 98.64 & 98.37 & 96.61 & 95.80 & \textbf{99.40} & 98.79 \\
agnews       & 69.95 & 96.93 & \textbf{97.82} & 94.08 & 89.67 & 92.25 & 90.97 & 90.60 & 96.20 \\
MultiRC      & 80.12 & 87.77 & 87.15 & 87.71 & 87.93 & 87.48 & \textbf{88.47} & 80.80 & 88.47 \\
yahoo        & \textbf{76.21} & 74.25 & 73.88 & 75.96 & 74.53 & 74.92 & 75.55 & 75.00 & 75.83 \\
\midrule
\rowcolor{gray!10}
\textbf{AP}  & 80.05 & 81.96 & 85.08 & 83.38 & 85.06 & 81.00 & 82.48 & \textbf{86.68} & 87.02 \\
\rowcolor{gray!10}
\textbf{BWT} & -4.46 & -5.52 & -1.77 & -2.62 & -0.43 & -5.82 & -4.28 & \textbf{-0.05} & -     \\
\bottomrule
\end{tabular}
}
\end{table}